\documentclass[nonatibib]{ifacconf}
\usepackage{subfig}
\usepackage{graphicx}      
\usepackage{natbib}        
\usepackage{bm}
\usepackage{amsmath,amssymb}
\usepackage{algpseudocode}
\usepackage{algorithm}
\algnewcommand\algorithmicinput{\textbf{Input:}}
\algnewcommand\Input{\item[\algorithmicinput]}
\usepackage{color}
\usepackage{booktabs}
\usepackage{tabularx}
\usepackage{caption}

\begin{document}
\begin{frontmatter}
\title{$\beta$-Variational Classifiers Under Attack} 

\thanks[footnoteinfo]{Part of this work was supported by MIUR (Italian Minister for Education)
under the initiative "Departments of Excellence" (Law 232/2016). We would like to also thank Nvidia for donating the Nvidia Titan V GPU used in for this research.}

\author[First]{Marco Maggipinto} 
\author[First]{Matteo Terzi} 
\author[Second]{Gian Antonio Susto}

\address[First]{Department of Information Engineering (DEI), University of Padova, Italy (e-mail: marco.maggipinto@phd.unipd.it, terzimat@dei.unipd.it))}
\address[Second]{DEI and Human-Inspired Technology Center, University of Padova, Italy (e-mail: gianantonio.susto@dei.unipd.it)}

\begin{abstract}                
Deep Neural networks have gained lots of attention in recent years thanks to the breakthroughs obtained in the field of Computer Vision. However, despite their popularity, it has been shown that they provide limited robustness in their predictions. In particular, it is possible to synthesise small adversarial perturbations that imperceptibly modify a correctly classified input data, making the network confidently misclassify it. This has led to a plethora of different methods to try to improve robustness or detect the presence of these perturbations. In this paper, we perform an analysis of $\beta$-Variational Classifiers, a particular class of methods that not only solve a specific classification task, but also provide a generative component that is able to generate new samples from the input distribution. More in details, we study their robustness and detection capabilities, together with some novel insights on the generative part of the model.
\end{abstract}

\begin{keyword}
Adversarial Training, Computer Vision, Deep Learning, Machine Learning, Robustness
\end{keyword}

\end{frontmatter}

\section{Introduction}
The astounding performance that Deep Neural Networks (DNNs) provide when dealing with large amounts of complex data has recently led to extensive research in Deep Learning (DL) technologies. Empirical evidence shows that, in contrast to standard Machine Learning (ML) methods, DNNs are able to generalize well in the over-parametrized regime (\cite{belkin2018understand, belkin2018does}), i.e. when the number of parameters of the model is much higher than the number of data used to train it; hence, there is basically no limit, other than the computational capabilities, to the complexity of the hypothesis class of functions that is of practical use. Despite this property, that is still not well understood and object of an entire line of research, the high complexity of the input-output relationship comes at a cost: the predictions provided by DNNs are not interpretable, making it difficult to understand what caused the model to take a particular decision; moreover, (\cite{szegedy2013intriguing}) discovered that DNNs are susceptible to adversarial perturbations, small changes in the input space that result in high changes in the output space. This allows the creation of \textit{Adversarial Examples} (\cite{goodfellow2014explaining}): for example, in image classification, it is possible to synthesise artificial images that, while visually identical for the human eye to a correctly classified sample, they are confidently misclassified.
While such problem is common in ML, it is emphasized in DNNs by the high dimensionality of the input space and the complexity of the function described by the DNN that may be subject to high curvature directions that can be exploited, even in a small neighborhood of a point, to significantly change the response of the network. 

The discovery started a completely new research trend that tries to understand the phenomenon (see \cite{liu2016delving, shaham2018understanding}) or find ways to defend against it. The most common approach to train robust networks is \textit{Adversarial Training} (\cite{goodfellow2014explaining,madry2017towards, terzi2019directional})  that consists in generating adversarial examples and feeding them to the network during training along with the correct label. While effective, such method comes at the cost of reduced prediction accuracy (\cite{tsipras2018robustness}) and increased training time compared to standard models.  Other approaches have been proposed to obtain robust models such as gradient regularization (\cite{ross2018improving}), Lipschitz regularization (\cite{finlay2018lipschitz}) and curvature regularization (\cite{moosavi2019robustness}). 
A different research line focuses on developing methods to detect adversarial examples, without requiring the model to be robust. In (\cite{feinman2017detecting}) it is proposed to combine Kernel Density Estimation on the hidden layer of the network (\cite{botev2010kernel}) and MC-Dropout (\cite{gal2016dropout}) that provides an estimate of the prediction uncertainty of the network. The rationale behind the method is that adversarial examples should have lower likelihood according to the estimated density and higher prediction uncertainty. 
(\cite{gong2017adversarial}) propose to train a binary classifier as a detection method: it is shown that such approach is able to detect 99\% of adversarial examples and it is robust to a second attack that aims at fooling the detection method.
(\cite{grosse2017statistical}) uses the kernel-based two-sample test (\cite{gretton2012kernel}) to detect statistical differences between adversarial examples and normal data. 

Recently, (\cite{li2018generative}) proposed a study on the robustness of Generative Classifiers and their detection capabilities. They propose three detection methods whose rejection policies are respectively: 1) reject samples with likelihood of the input lower than a certain threshold; 2) Reject samples with joint input/output likelihood lower than a certain threshold; 3) Reject over/under confident predictions. The methods have proven effective on object recognition tasks. Moreover, they show that models with lower capacity are more robust to adversarial examples.

We build upon (\cite{li2018generative}) to provide an analysis of $\beta$-Variational Classifiers, a similar approach but based on $\beta$-Variational Autoencoders (\cite{higgins2017beta}) that are able to provide a disentangled representation of the input (\cite{burgess2018understanding}) and give an alternative method to control the model capacity. In particular, the contributions of our paper are as follows:
\begin{itemize}
    \item We analyze the robustness of $\beta$-Variational Classifiers combined with sparse regularization;
    \item We analyze the detection capabilities of $\beta$-Variational Classifiers;
    \item We analyse the effects of adversarial perturbations on the decoder network.
\end{itemize}
The remainder of this paper is organized as follows: In section \ref{vae} and \ref{vac} we provide a description of $\beta$-Variational Autoencoders and their  $\beta$-Variational Classifiers. In Section \ref{adv} we explain the phenomenon of adversarial examples and the main method used to synthesise them. In Section \ref{set} and \ref{res} we describe the experimental settings and outline the obtained results. Finally, in Section \ref{concl} conclusions and future works are reported.

\section{$\beta$-Variational-Autoencoders} \label{vae}
Generative modeling aims at learning a parametrized model of the probability distribution underlying the data in order to obtain new realistic samples from it. In this context, Variational Autoencoders (VAEs) (\cite{kingma2013auto}) are a well know approach to develop a complex latent variable model that can be learned by Stochastic Gradient Descent (SDG).\\
\begin{figure}
    \centering
    \includegraphics[width=\columnwidth]{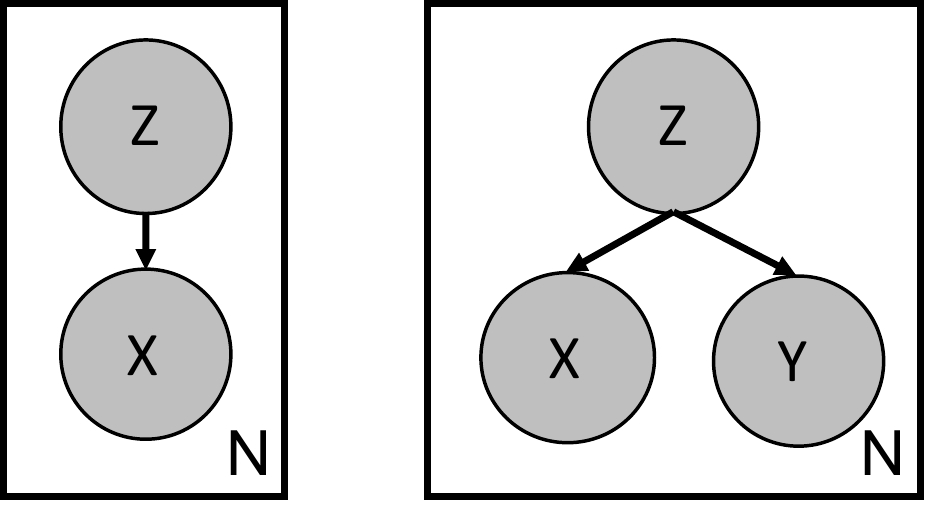}
    \caption{Bayesian network of a VAE (left) and the Variational Classifier analyzed in this work (right).}
    \label{fig:bnet}
\end{figure}
Given a dataset of independent identically distributed samples $\{ \bm{x}_i \}$ $i = 1, \cdots, N$ with distribution $p(\bm{x})$, we introduce hidden variables $\{ \bm{z}_i \}$ $i = 1, \cdots, N$ of dimension $d$ distributed as a multivariate Gaussian $p(\bm{z}) \sim \mathcal{N}(\bm{0}, \bm{I})$ with $\bm{I} \in \mathcal{R}^{d \times d}$ the identity matrix. \\
VAEs model the joint distribution of the random variables $\bm{X}$ and $\bm{Z}$ as $p_{\theta}(\bm{x}, \bm{z}) = p_{\bm{\theta}}(\bm{x} | \bm{z}) p(\bm{z})$ corresponding to the bayesian network in Figure \ref{fig:bnet}(left) where the mean of  $p_{\bm{\theta}}(\bm{x} | \bm{z})$ is the output of a Decoder Neural Network $D_{\bm{\theta}}(\bm{z})$ parametrized by parameters $\bm{\theta}$, typical choices are Gaussian $p_{\bm{\theta}}(\bm{x} | \bm{z}) \sim \mathcal{N}(D_{\bm{\theta}}(\bm{z}), \bm{I})$ for continuous output or Bernoulli $p_{\bm{\theta}}(\bm{x} | \bm{z}) \sim \mathcal{B}(D_{\bm{\theta}}(\bm{z}))$ for binary output. Being the $\bm{z}_i$ unknown, we aim at finding the parameters value that maximizes the marginal likelihood; however, this requires evaluating an intractable integral to compute $p_{\bm{\theta}}(\bm{x}) = \mathbb{E}_{\bm{z}} \left[ p_{\bm{\theta}}(\bm{x} | \bm{z}) \right]$, which is also difficult to approximate by means of Monte Carlo methods due to the dimensionality of the hidden factors and the amount of data that is often very high in Deep Learning settings. In a similar scenario approximate inference is typically very effective; more in details, introducing an approximate posterior distribution $q_{\bm{\phi}}(\bm{z} | \bm{x}) \sim \mathcal{N}(\bm{\mu}(\bm{x}, \bm{\Sigma}(\bm{x}))$ where $\bm{\mu}(\bm{x}), \bm{\Sigma}(\bm{x})$ are output of an Encoder network. We define the Expectation Lower bound (ELBO) $\mathcal{L}(\bm{\theta}, \bm{\phi}, \bm{x})$ as:
\begin{equation} \label{elbo}
\begin{split}
    \mathcal{L}(\bm{\theta}, \bm{\phi}, \bm{x}) =  & -D_{KL}\left(q_{\bm{\phi}}(\bm{z}|\bm{x}) \, || \, p(\bm{z})\right) + \\ & \mathbb{E}_{q_{\bm{\phi}}}(\bm{z} | \bm{x}) \left[ \log \, p_{\bm{\theta}}(\bm{x} | \bm{z}) \right]
\end{split}
\end{equation}
Where $D_{KL}$ is the Kullback–Leibler divergence. It is always true that (for a detailed proof see \cite{bishop2006pattern}):
\begin{equation} \label{eq:bound}
    	\log\left(p_{\bm{\theta}}(\bm{x})\right) \geq \mathcal{L}(\bm{\theta}, \bm{\phi}, \bm{x})
\end{equation}
(\ref{eq:bound}) has important implications when the parametrized distributions are extremely complex, such as Neural Networks. We can in fact maximize the ELBO instead of the intractable marginal likelihood. In particular, if we analyze the expression in (\ref{elbo}), the Decoder network is trained to minimize an expected reconstruction error, while the Encoder network distribution is pushed to be close to the prior, this acts as a regularizer, tuning the capacity of the Encoder. Such regularizer impacts the type of representations that the Encoder can learn, in particular (\cite{belkin2018understand}) showed that by controlling this term using the following modified ELBO:
\begin{equation} \label{beta:elbo}
\begin{split}
    \mathcal{L}(\bm{\theta}, \bm{\phi}, \bm{x}) =  & \beta | D_{KL}\left(q_{\bm{\phi}}(\bm{z}|\bm{x} \, || \, p(\bm{z})\right) - C |+ \\ & \mathbb{E}_{q_{\bm{\phi}}}(\bm{z} | \bm{x}) \left[ \log \, p_{\bm{\theta}}(\bm{x} | \bm{z}) \right]
\end{split}
\end{equation}
The model is able to produce disentangled representations that are related to different characteristics of the image, e.g. color, shape etc. Here $\beta$ and $C$ are hyperparameters, the first one is usually kept very high around 1000 while the second directly control the capacity and is linearly increased at training time from $0$ to a predefined value (this procedure has been shown to provide better representations). This modified version, is called $\beta$-VAE.

During optimization, $\mathbb{E}_{q_{\bm{\phi}}}(\bm{z} | \bm{x}) \left[ \log \, p_{\bm{\theta}}(\bm{x} | \bm{z}) \right]$ is computed using a Monte Carlo approximation  $\mathbb{E}_{q_{\bm{\phi}}}(\bm{z} | \bm{x}) \left[ \log \, p_{\bm{\theta}}(\bm{x} | \bm{z}) \right]$ $\approx \frac{1}{M} \sum_{m=1}^M log \, p_{\bm{\theta}}(\bm{x} | \bm{z}^{(m)})$ $\{ \bm{z}_{m} \}_{m=1}^M \sim q_{\bm{\phi}}(\bm{z} | \bm{x})$. Since it is not possible to back-propagate through the sampling operation, a reparametrization trick is used i.e. $\bm{z}_{m} = \mu(\bm{x}) + \xi_{m} \bm{\Sigma}(\bm{x})$ with $\xi_{m} \sim \mathcal{N}(0, 1)$.
The detailed optimization procedure is reported in Algorithm \ref{alg:vae_tr}.

\begin{algorithm}[!h] 
\caption{Training Procedure}\label{alg:vae_tr}
\begin{algorithmic}[1] 
\Input $\{\bm{x}^{(i)}\}_{i=1}^n, B, M, \beta, C, n_{iter}$
\For{$j = 1 \dots n_{iter}$}
\State Sample $B$ examples from the training set $\{\bm{x}^{ \left( i \right) }\}_{i=1}^B$
		\State Sample $B \cdot M$ latent variables $\bm{z}^{(i)}_m$
		\State $c = \text{linearSchedule}(C, j)$
		\State Compute the gradient with respect to $\bm{\Phi}, \bm{\theta}$
		\begin{equation*}
		\bm{\delta}_{\bm{\theta}} = \nabla_{\bm{\theta}} \frac{1}{B \cdot M}  \sum_{i=1}^{B} \sum_{m=1}^M \log \, p_{\bm{\theta}}(\bm{x}^{(i)} | \bm{z}^{(i)}_m) 
		\end{equation*}
				\begin{equation*}
				\begin{split}
			\bm{\delta}_{\bm{\Phi}} = &  \nabla_{\bm{\Phi}} \frac{1}{B} \sum_{i=1}^{B}  \left[ \beta \, | D_{KL}\left(q_{\bm{\phi}}(\bm{z}|\bm{x}^{(i)}) \, || \, p(\bm{z})\right) - c | \, + \right. \\ & \left. \, + \frac{1}{M}\sum_{m=1}^M \log \, p_{\bm{\theta}}(\bm{x}^{(i)} | \bm{z}^{(i)}_m) \right]
				\end{split}
		\end{equation*}
		\State Ascend the gradient $\bm{\theta} = \text{ascendRule} \left( \bm{\theta}, \bm{\delta}_{\theta}\right) $
		\State Ascend the gradient $\bm{\Phi} = \text{ascendRule} \left( \bm{\theta}, \bm{\delta}_{\Phi}\right) $
\EndFor 
\end{algorithmic}
\end{algorithm}

\section{$\beta$-Variational-Classifiers} \label{vac}
$\beta$-VAEs provide an interesting method to perform variational inference in the presence of complex parametrized models of probability distributions with hidden factors. A similar optimization procedure can be employed to learn a more complex Bayesian model that includes a random variable $\bm{Y}$ representing a class which the input belongs to. The  resulting model is called  $\beta$-Variational Classifier ($\beta$-VAC) . In this work, we focus our study on a particular $\beta$-VAC  represented by the Bayesian network in Figure \ref{fig:bnet} (right), we assume that the dataset is composed by couples $\{\bm{x}^{(i)}, y^{(i)}\}_{i=1}^{n}$ where $y^{(i)}$ are the true labels (i.e. object classes) associated to the input. Introducing the conditional distribution $p_{\bm{\omega}}(y | \bm{z})$. The ELBO for such model is: 
\begin{equation}
\begin{split}
    	\mathcal{L}(\bm{\theta}, \bm{\phi}, \bm{x}, y) = & -D_{KL}\left(q_{\bm{\phi}}(\bm{z}|\bm{x}, y) \, || \, p(\bm{z})\right) + \\
    	& + \mathbb{E}_{q_{\bm{\phi}}(\bm{z} | \bm{x}, y)} \left[ \log \, p_{\bm{\theta}}(\bm{x} | \bm{z})  p_{\bm{w}}(y | \bm{z}) \right]
\end{split}
\end{equation}
From now on, we assume that $q_{\bm{\phi}}(\bm{z}|\bm{x}, y) = q_{\bm{\phi}}(\bm{z}|\bm{x})$ which states that all the information about $\bm{z}$ is contained in $\bm{x}$. The resulting ELBO, including the modified regularizer of Section \ref{vae} can be expressed as follows:

\begin{equation}
\begin{split}
    	\mathcal{L}(\bm{\theta}, \bm{\phi}, \bm{x}, y) = & \beta | D_{KL}\left(q_{\bm{\phi}}(\bm{z}|\bm{x}) \, || \, p(\bm{z})\right) - C | + \\
    	& + \mathbb{E}_{q_{\bm{\phi}}(\bm{z} | \bm{x})} \left[ \log \, p_{\bm{\theta}}(\bm{x} | \bm{z})  p_{\bm{w}}(y | \bm{z}) \right]
\end{split}
\end{equation}
The optimization procedure to learn the model parameters is analogous to Algorithm \ref{alg:vae_tr}.


\section{Adversarial Examples} \label{adv}
\begin{figure}
    \centering
    \includegraphics[width=1.0\columnwidth]{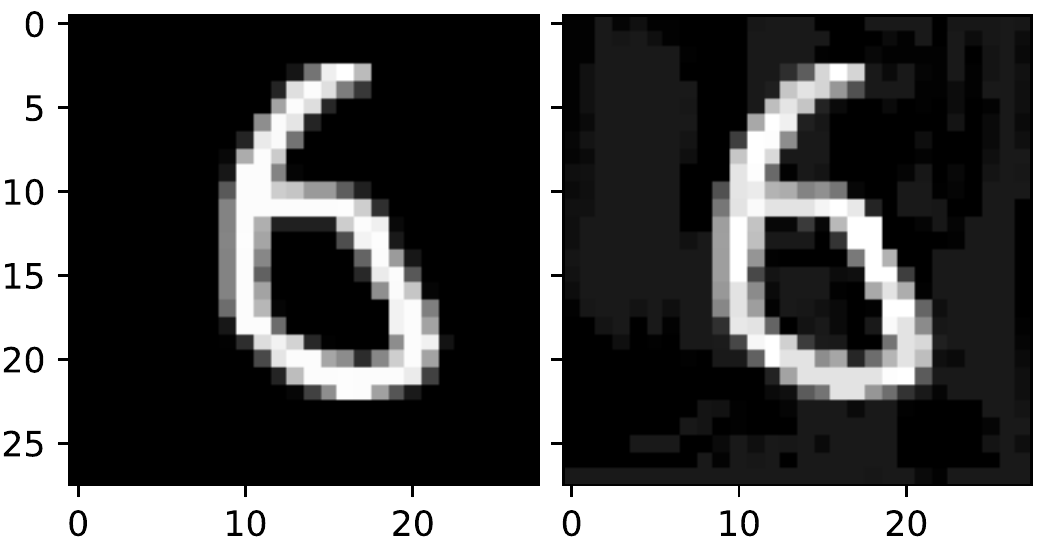}
    \caption{Adversarial example on a digit recognition task: the 6 on the left is correctly classified while the one on the right is classified a 4.}
    \label{fig:adv_ex}
\end{figure}

In object recognition, an adversarial example is an image that, while being visually indistinguishable or very similar to a correctly classified input, is confidently mis-classified by the model. The most common approach to synthetize an adversarial example $\bm{x}_{adv}$ is the Projected Gradient Descent (PGD) attack, where a normal data $\bm{x}$ is perturbed by following an ascending direction of the loss function while remaining in an $\epsilon$-ball $B_{\epsilon}(\bm{x}) = \{ \bm{x}_{adv} \, s.t. \,||\bm{x}-\bm{x}_{adv}||_p \leq \epsilon\}$ centered at the original sample. More in details, let $L(\bm{x}, y$) be the value of the loss at $\bm{x}$ where $y$ is the correct label, $\epsilon > 0$ the maximum distance from the original input relative to the maximum input value (e.g. 255 for 8bit images), $k$ the number of iterations and $\alpha$ the step size, PGD works as desctibed in Algorithm \ref{alg:pgd}.
\begin{algorithm}[!h] 
\caption{PGD}\label{alg:pgd}
\begin{algorithmic}[1] 
\Input $\bm{x}, y,  k, \epsilon, \alpha$
\State $\bm{x}_0 = x$
\For{$j = 1 \dots k$}
	\State $\bm{x}_j = \bm{x}_j \,  +  \, \alpha \nabla L(\bm{x}_j, y)$
	\State $\bm{x}_j = Proj(\bm{x}_j, B_\epsilon(\bm{x}))$
\EndFor 
\State $\bm{x}_{adv} = \bm{x}_j$ \\
\Return $\bm{x}_{adv}$
\end{algorithmic}
\end{algorithm}
\\$Proj(\bm{x}_j, B_\epsilon(\bm{x}))$ is the projection operator that varies depending on the chosen norm, typical choices are $\ell_2$ and $\ell_\infty$. The value of $\epsilon$ identifies the strength of the applied perturbation. Typically, for simple tasks such as handwritten digit recognition, an high value is necessary to fool the network while for more complex tasks, the value can be much smaller. Figure \ref{fig:adv_ex} shows an adversarial example on a digit recognition task dataset, with $\epsilon = 0.1$.

\section{Experimental settings} \label{set}
In order to analyze the robustness of $\beta$-VAC to adversarial examples we train the model on two popular datasets in Computer Vision for handwritten digit classification\footnote{http://yann.lecun.com/exdb/mnist} (MNIST) and for clothes recognition\footnote{https://github.com/zalandoresearch/fashion-mnist} (Fashion-MNIST or FMNIST by Zalando research). 
They both are composed by 60 thousand labeled images for the train set and 10 thousand for testing, the images are in grayscale of size $28\times28$.
 
The structure of the model is as follows:
\begin{itemize}
    \item Encoder: is an adapted Allcnn (\cite{springenberg2014striving}) to provide an hidden representation of size 100 i.e. $\bm{z} \in \mathbb{R}^{100}$;
    \item Decoder: has a structure symmetric to the encoder, in order to provide as output an image of the same dimension of the input;
    \item Classifier: is a 2 layer perceptron with 64 hidden neurons per layer and ReLu (\cite{nair2010rectified}) activations.
\end{itemize}
We train for 60 epochs using an SGD optimizer with momentum 0.9 and learning rate 0.01 that is decreased at the 10th and 30th epoch by a factor of 10. The strength of the momentum is a common choice in literature while the other attributes have been chosen in order to properly train the network. We employ a small weight decay of $\text{1e-6}$ (typical values are around 1e-3 for the Allcnn) so its influence on the capacity is limited and we are free to control it using the KL regularization term in the ELBO. We use PGD to compute the adversarial examples with variable strength, in particular for MNIST we use $\epsilon \in [0, 0.3]$ while for FMNIST $\epsilon \in [0, 0.1]$; the choice is justified by the more difficult classification task of the second dataset so a smaller perturbation is sufficient to effectively attack the network. We keep the number of iterations equal to 40 and the step size 0.01, the value of $\epsilon$ is referred to the $\ell_{\infty}$ norm.

\section{Results} \label{res}
In this section we outline the obtained results and, in particular, we analyze the effect of the capacity and sparse regularization on the robustness and detection capabilities of the $\beta$-VAC described in the previous section. To conclude, we visually inspect the reconstructed images of adversarial examples, showing some interesting properties.

  \begin{figure*}[!h]
    	\subfloat[Reconstruction error]{ 
    	 	\includegraphics[width=0.33\textwidth, height=4.0cm]{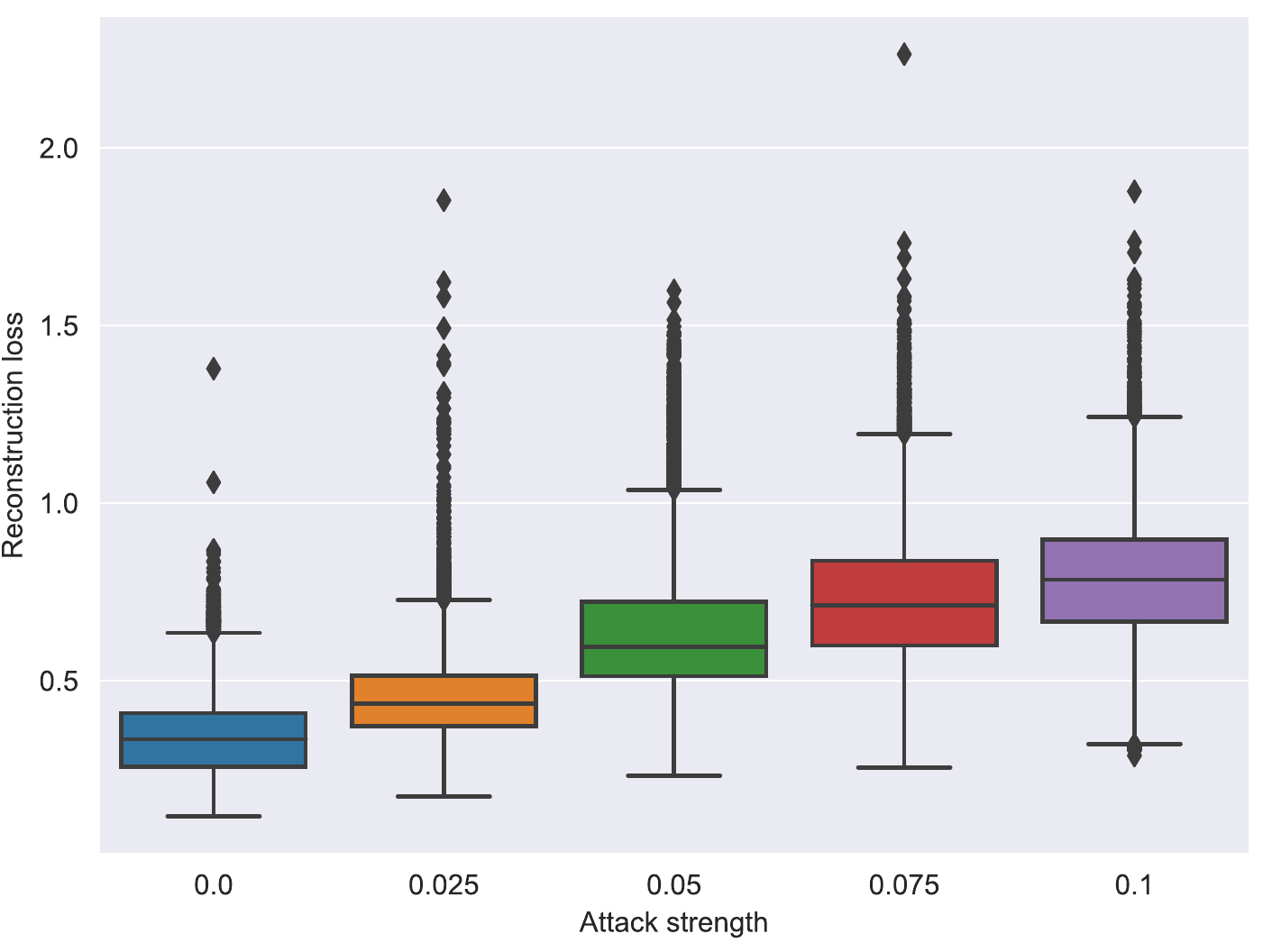}}
     \subfloat[Model accuracy]{ 
    	 	\includegraphics[width=0.33\textwidth, height=4.0cm]{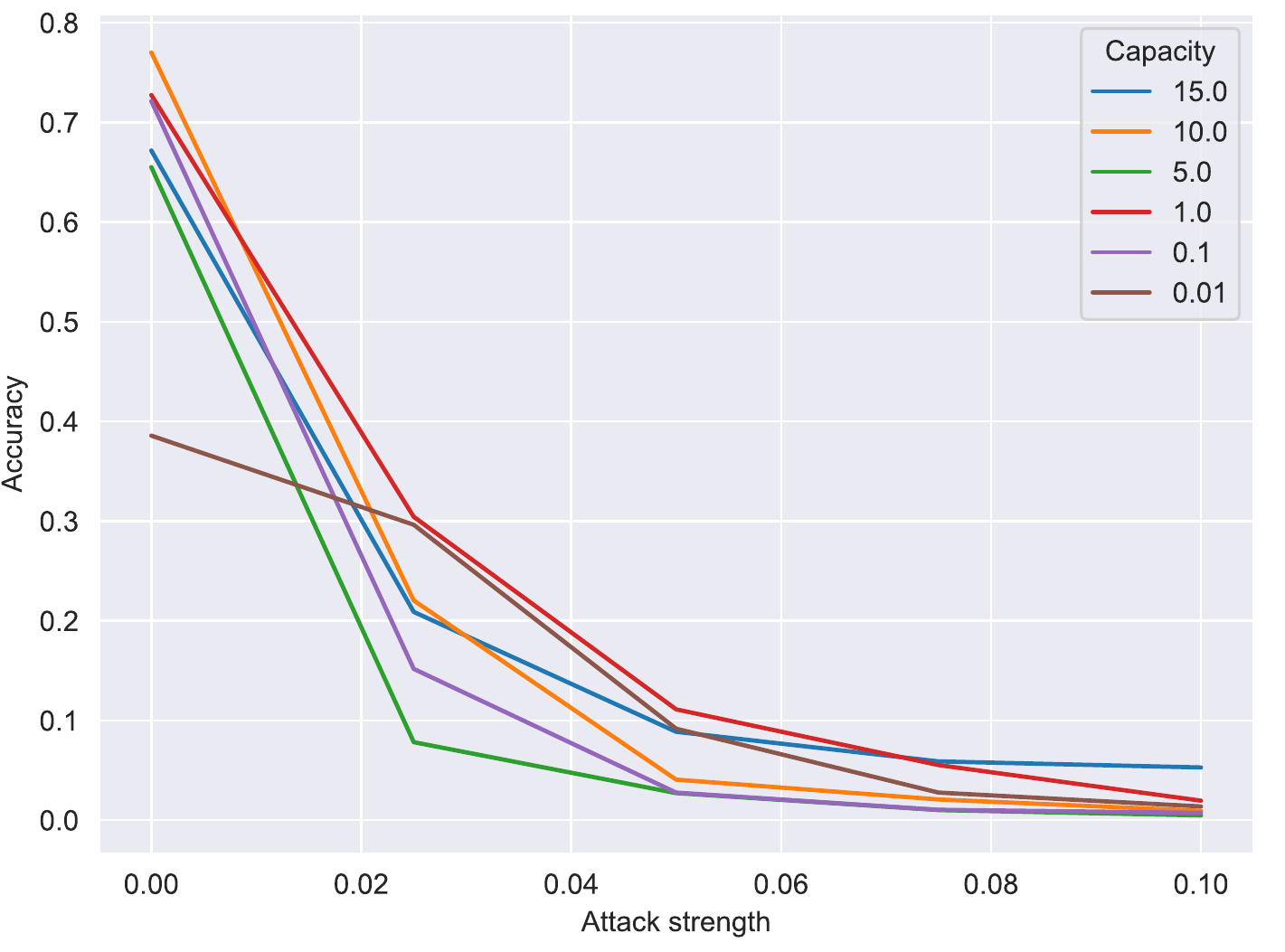}}
    	\subfloat[Detection rate]{ 
    	 	\includegraphics[width=0.33\textwidth, height=4.0cm]{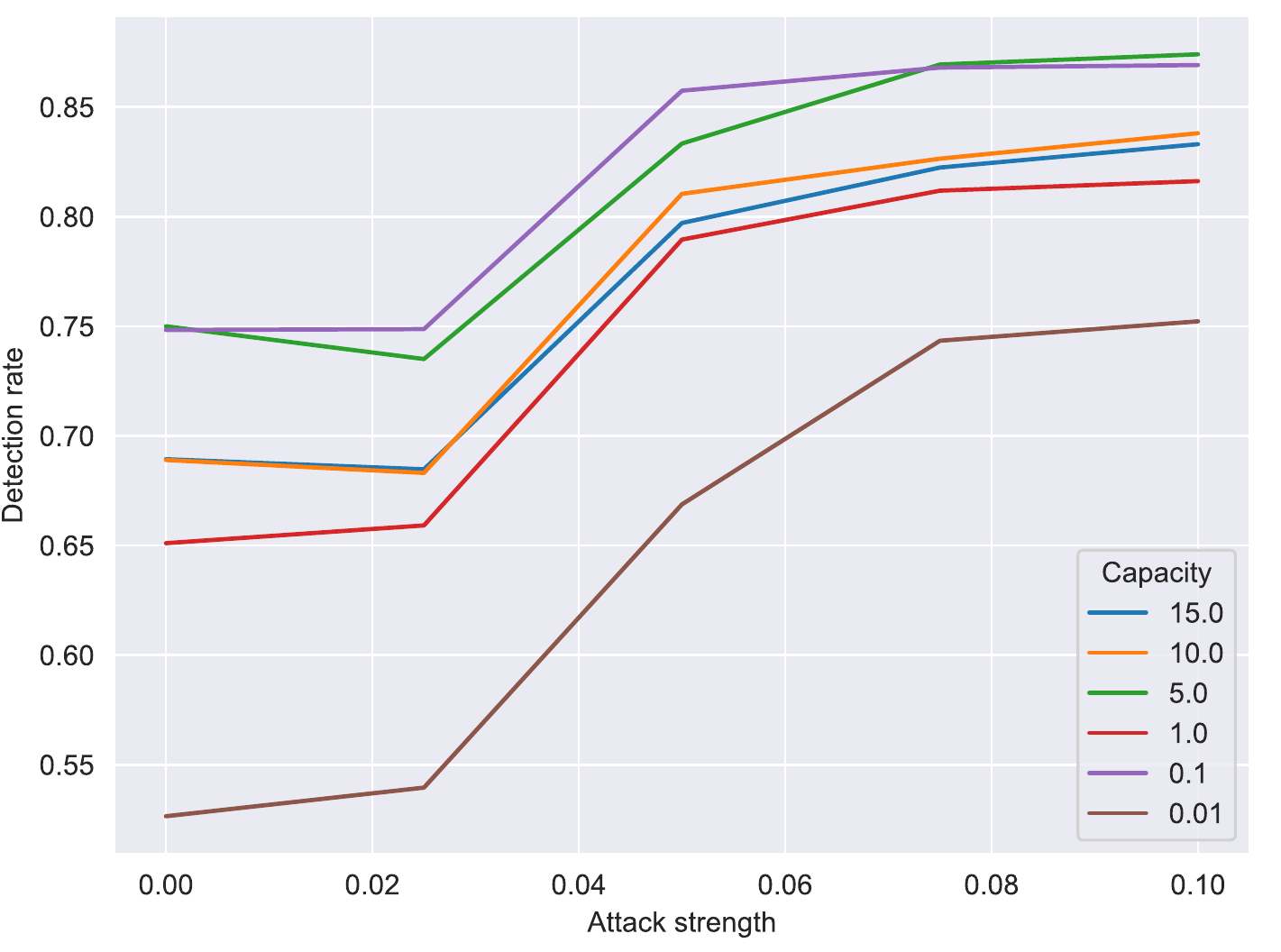}}
   \caption{Results for the FMNIST dataset.}
  \label{fig:mnist} 
  \end{figure*}
  
   \begin{figure*}[!h]
    	\subfloat[Reconstruction error]{ 
        		 \includegraphics[width=0.33\textwidth, height=4.0cm]{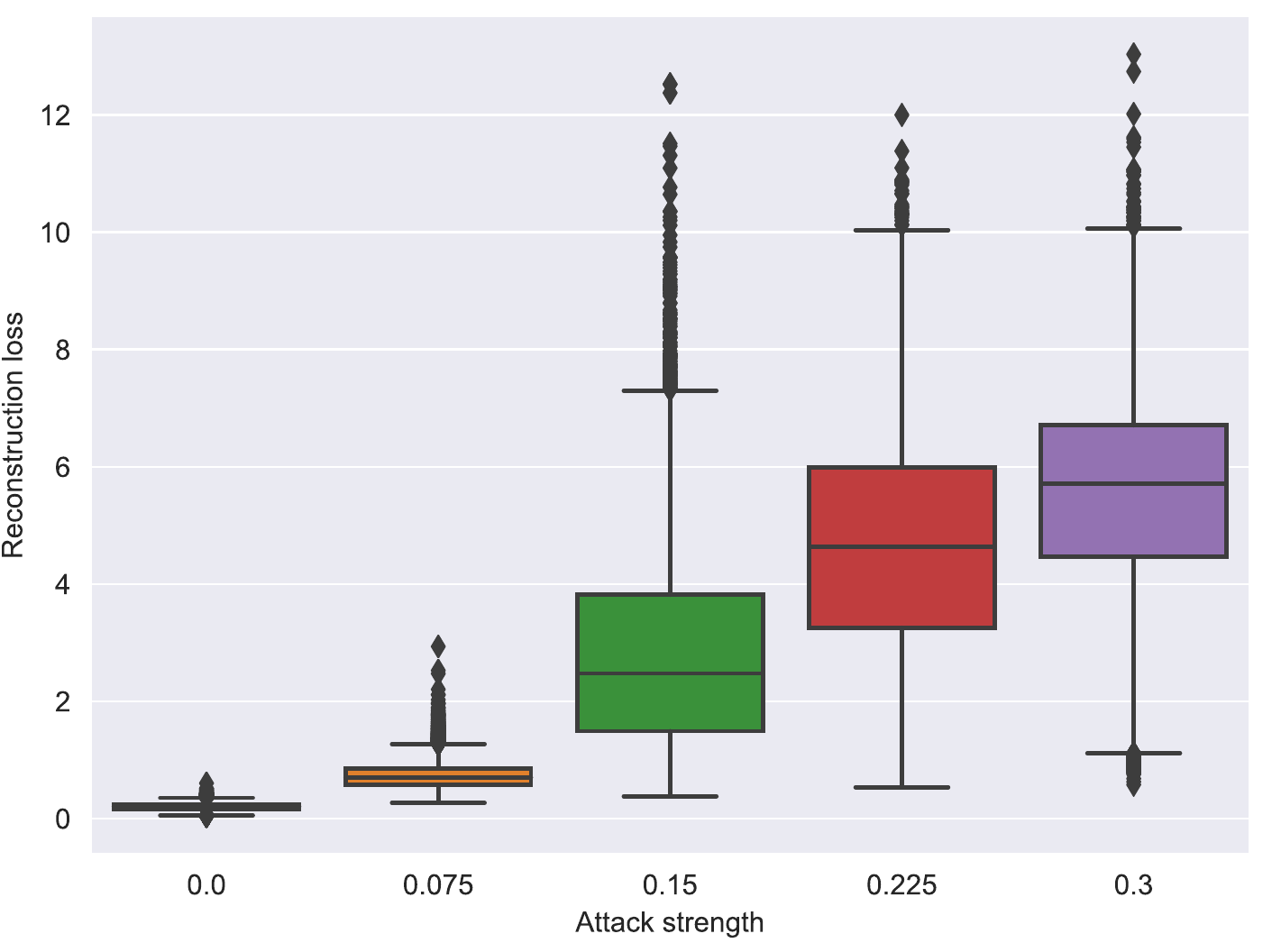}} 
    	   	\subfloat[Model accuracy]{ 
        		 \includegraphics[width=0.33\textwidth, height=4.0cm]{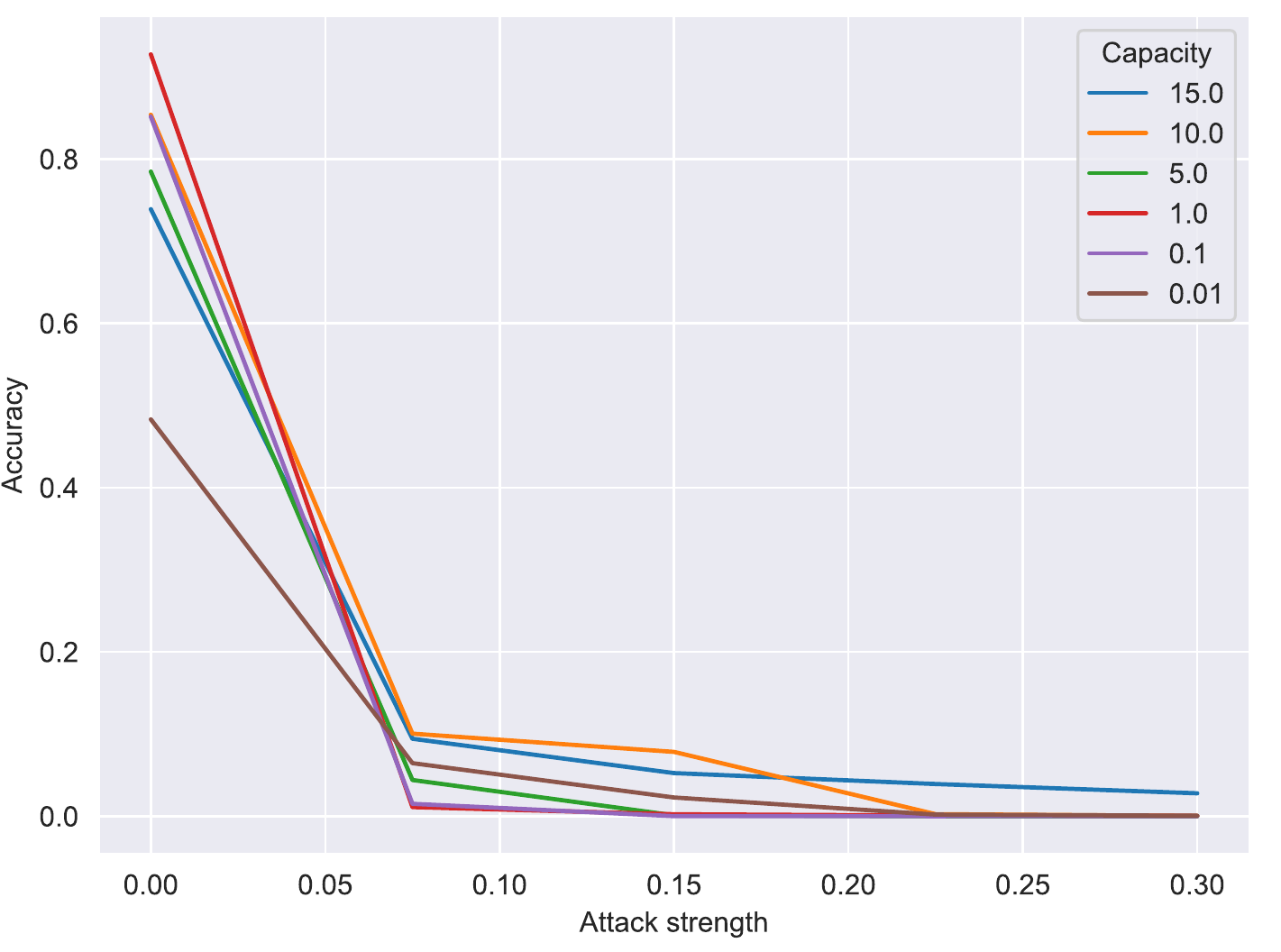}} 
       	\subfloat[Detection rate]{ 
        		 \includegraphics[width=0.33\textwidth, height=4.0cm]{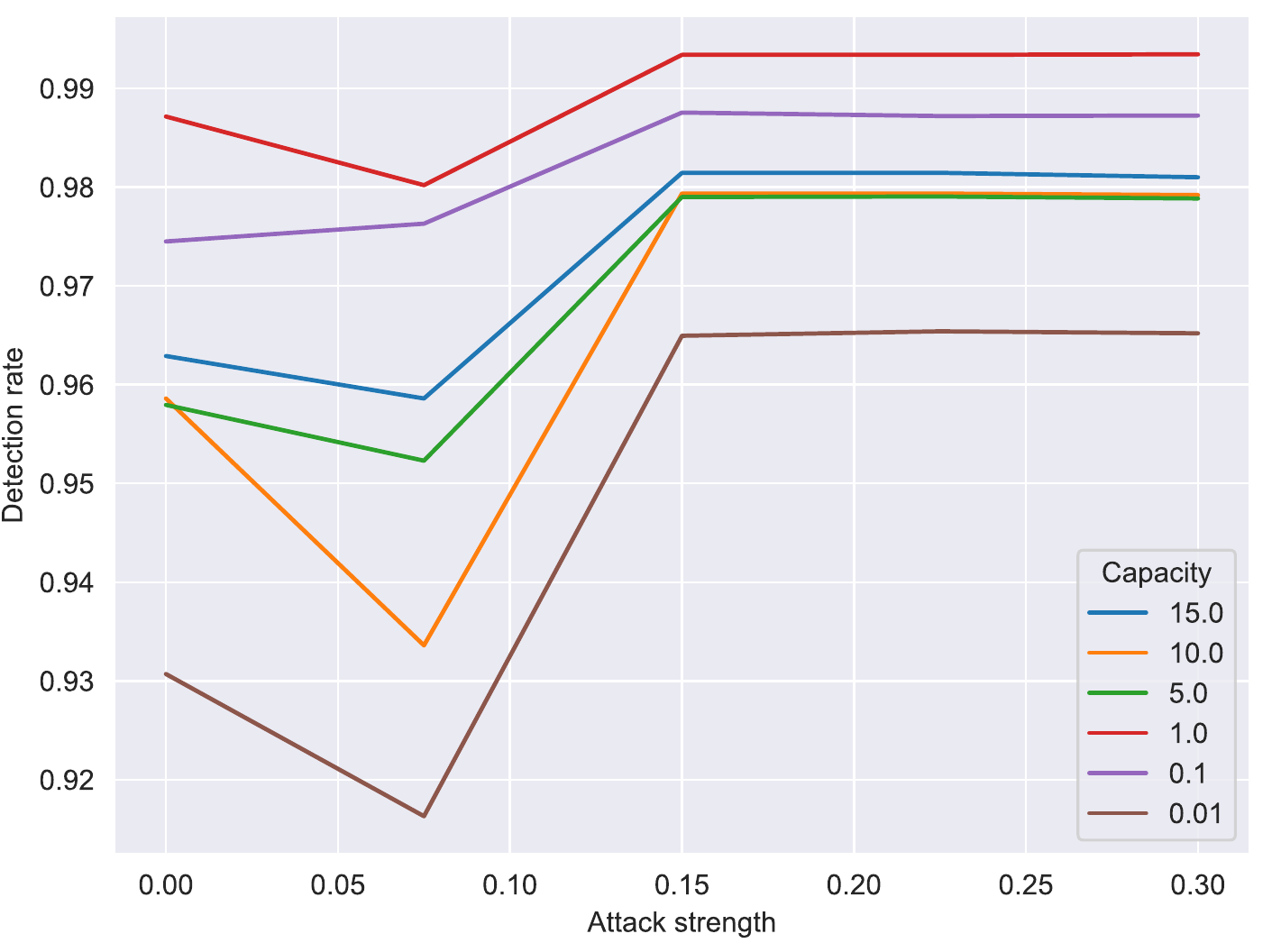}} 
   \caption{Results for the MNIST dataset.}
  \label{fig:fmnist} 
  \end{figure*}

 \begin{table*}[]
 \captionsetup{width=1.0\linewidth}
 \caption{$l1$ regularization effect on the robustness for the MNIST dataset.}
 \begin{tabularx}{1.0\textwidth}{XXXXXX}
\toprule
{} &    $\epsilon = 0.0$ &     $\epsilon =0.075$ &     $\epsilon = 0.150$ &     $\epsilon =0.225$ &     $\epsilon = 0.3$ \\
\midrule
1e-06 &  0.9395 &  0.1360 &  0.0988 &  0.0918 &  0.0471 \\
5-07 &  0.8461 &  0.0111 &  0.0004 &  0.0002 &  0.0001 \\
1-07 &  0.8452 &  0.0498 &  0.0218 &  0.0215 &  0.0254 \\
5e-08 &  0.8320 &  0.1194 &  0.0646 &  0.0527 &  0.0377 \\
\bottomrule
\end{tabularx}
\label{tab:l1mnistrob}
\end{table*}

 \begin{table*}[]
 \captionsetup{width=1.0\linewidth}
 \caption{$l1$ regularization effect on the robustness for the FMNIST dataset.}
 \begin{tabularx}{1.0\textwidth}{XXXXXX}
\toprule
{} &   $\epsilon = 0.0$ &     $\epsilon = 0.025$ &     $\epsilon = 0.050$ &     $\epsilon = 0.075$ &     $\epsilon = 0.1$  \\
\midrule
1e-03 &  0.624 &  0.185 &  0.083 &  0.027 &  0.009 \\
1e-04 &  0.657 &  0.081 &  0.041 &  0.026 &  0.020 \\
1e-05 &  0.722 &  0.260 &  0.140 &  0.089 &  0.063 \\
1e-06 &  0.764 &  0.216 &  0.084 &  0.048 &  0.033 \\
\bottomrule
\end{tabularx}
\label{tab:l1fmnistrob}
\end{table*}

 \begin{table*}[]
 \captionsetup{width=1.0\linewidth}
 \caption{$\ell_1$ regularization effect on the detection accuracy for the MNIST dataset.}
 \begin{tabularx}{1.0\textwidth}{XXXXXX}
 \toprule
 { $\ell_1$ strength} &    $\epsilon = 0.0$ &     $\epsilon =0.075$ &     $\epsilon = 0.150$ &     $\epsilon =0.225$ &     $\epsilon = 0.3$ \\
 \midrule
 1e-06 &  0.976 &  0.976 &  0.988 &  0.988 &  0.988 \\
 5e-07 &  0.967 &  0.976 &  0.983 &  0.983 &  0.983 \\
 1e-07 &  0.968 &  0.969 &  0.984 &  0.984 &  0.984\\
5e-08&  0.945 &  0.953 &  0.972 &  0.972 &  0.972 \\
 \bottomrule
 \end{tabularx}
 \label{tab:l1mnistdet}
 \end{table*}
 
 \begin{table*}[]
 \captionsetup{width=1.0\linewidth}
 \caption{$\ell_1$ regularization effect on the detection accuracy for the FMNIST dataset.}
 \begin{tabularx}{1.0\textwidth}{XXXXXX}
 \toprule
 {$\ell_1$ strength} &     $\epsilon = 0.0$ &     $\epsilon = 0.025$ &     $\epsilon = 0.050$ &     $\epsilon = 0.075$ &     $\epsilon = 0.1$ \\
 \midrule
 1e-03 &  0.628 &  0.637 &  0.741 &  0.782 &  0.800 \\
 1e-04 &  0.738 &  0.716 &  0.812 &  0.843 &  0.852 \\
 1e-05 &  0.699 &  0.686 &  0.810 &  0.838 &  0.841 \\
 1e-06 &  0.713 &  0.720 &  0.830 &  0.846 &  0.850 \\
 \bottomrule
 \end{tabularx}
 \label{tab:l1fmnistdet}
 \end{table*}
 
\subsection{Adversarial detection}
The decoder part of the model can be extremely useful to detect adversarial examples: by definition, they provide an high variation of the network output given small variations of the input; hence, if the reconstructed image is strongly affected by the adversarial perturbation, it will probably be very different to the input fed to the network. An example of this phenomenon is shown in Figure \ref{fig:mnist}(a) and \ref{fig:fmnist}(a), it is noticeable how the the reconstruction error considerably increases at the strengthening of the attack. We can thus train a classifier on the reconstruction error to obtain an effective attack detection methods. In the following, we will use a logistic classifier trained on adversarial examples computed on the training set and we analyse its performance by computing adversarial examples on the test set, and classifying both clean input and perturbed ones. We use the classification rate as performance metric being the dataset well balanced due to the high effectiveness of adversarial attacks.
\subsection{Capacity effect}
In Figure \ref{fig:mnist}(b) and \ref{fig:fmnist}(b) we report the robustness  of the $\beta$-VAC, measured in terms of accuracy at classifying adversarial examples, as a function of the parameter $C$ controlling the capacity of the encoder network. We don't notice particular correlation between the capacity and the accuracy. On the FMNIST dataset, for small attack strength, there are some values of $C$ that provide better robustness i.e. 0.01 and 1.0, however the first one has low accuracy on clean samples ($\epsilon=0$) due to the limited capacity of the model. In general, we cannot conclude to be able to control the robustness by changing the capacity.

For the detection rate Figure \ref{fig:mnist}(c) and \ref{fig:fmnist}(c), the capacity seems to have a more strong effect, in particular when values are too low, such as 0.01, the reconstruction error is high also for normal samples, making it difficult to distinguish them from the adversarial ones. On the other hand, we don't see a positive correlation between capacity and detection rate, so the parameter has to be fine tuned to get the best result.
Overall, the detection accuracy is very good on the MNIST dataset also due to the higher attack strength used, for FMNIST the detection rate is well above 70\% with the right capacity for every attack strength.
\subsection{Sparse regularization effect}
The idea of adding sparse regularization is motivated by the fact that the encoder network of $\beta$-VAE has been shown to provide disentangled representations. Hence, if the classifier selects only the ones that are useful for the classification task and discard the others, it should be more robust to adversarial perturbations. In Table \ref{tab:l1mnistrob} and \ref{tab:l1fmnistrob} we show the effect of the $\ell_1$ regularization on the robustness of the classifier trained using the best performing capacity, i.e 1.0 for MNIST and 10.0 for FMNIST. 
The $\ell_1$ regularization does not seem to provide improved robustness. This result provides us the following insight: while it is easy to see that for linear models, obtaining $l_\infty$-robustness is equivalent to applying $\ell_1$-regularization, for non-linear model this relation is not true. Thus, our results give the evidence that the Encoder network is not providing robust latent embeddings.

Overall, the detection accuracy is lower than the one for the non regularized method. This would be a fair price to pay in case of improved robustness but from the results obtained it is probably better to avoid using $\ell_1$ regularization and choose the correct model complexity with $C$.

\subsection{Decoded images inspection}
\begin{figure}[!h]
  	\subfloat{%
      		 \includegraphics[width=1.0\columnwidth]{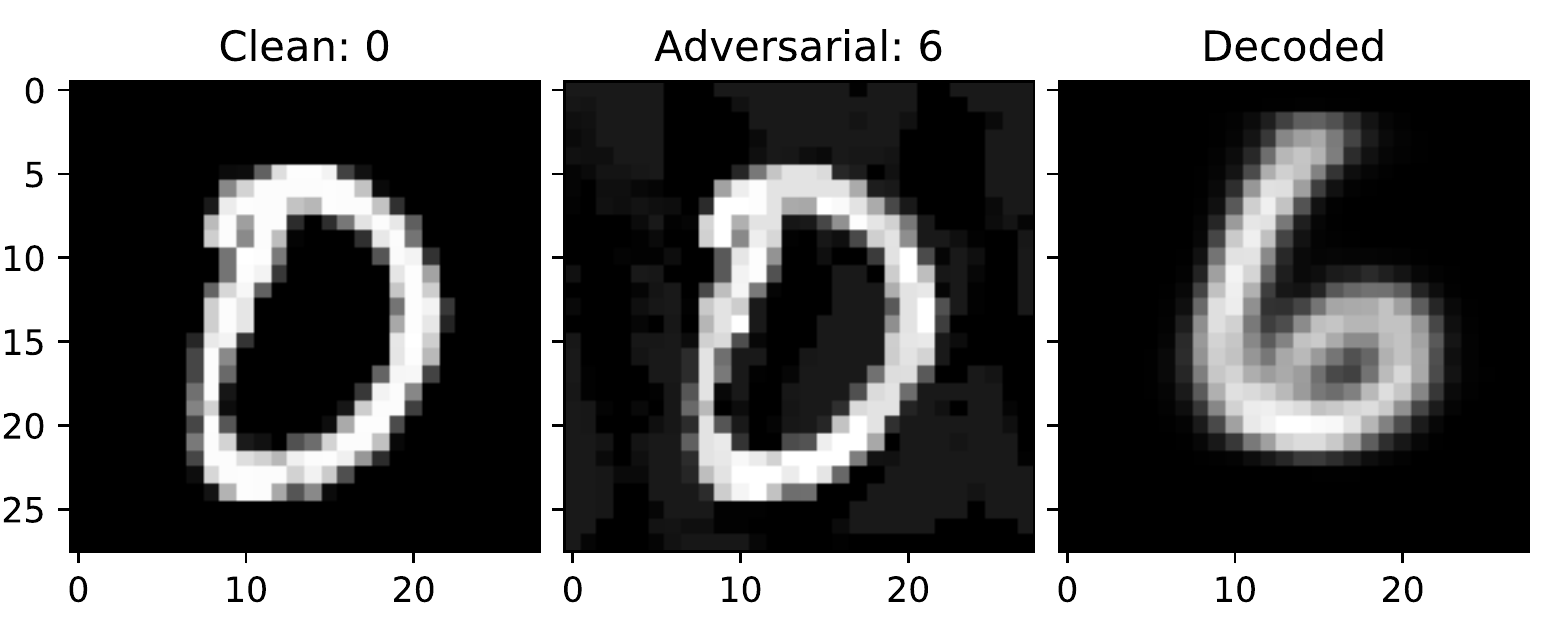}} 
      		 
  	 	
  	\subfloat{%
  	 	\includegraphics[width=1.0\columnwidth]{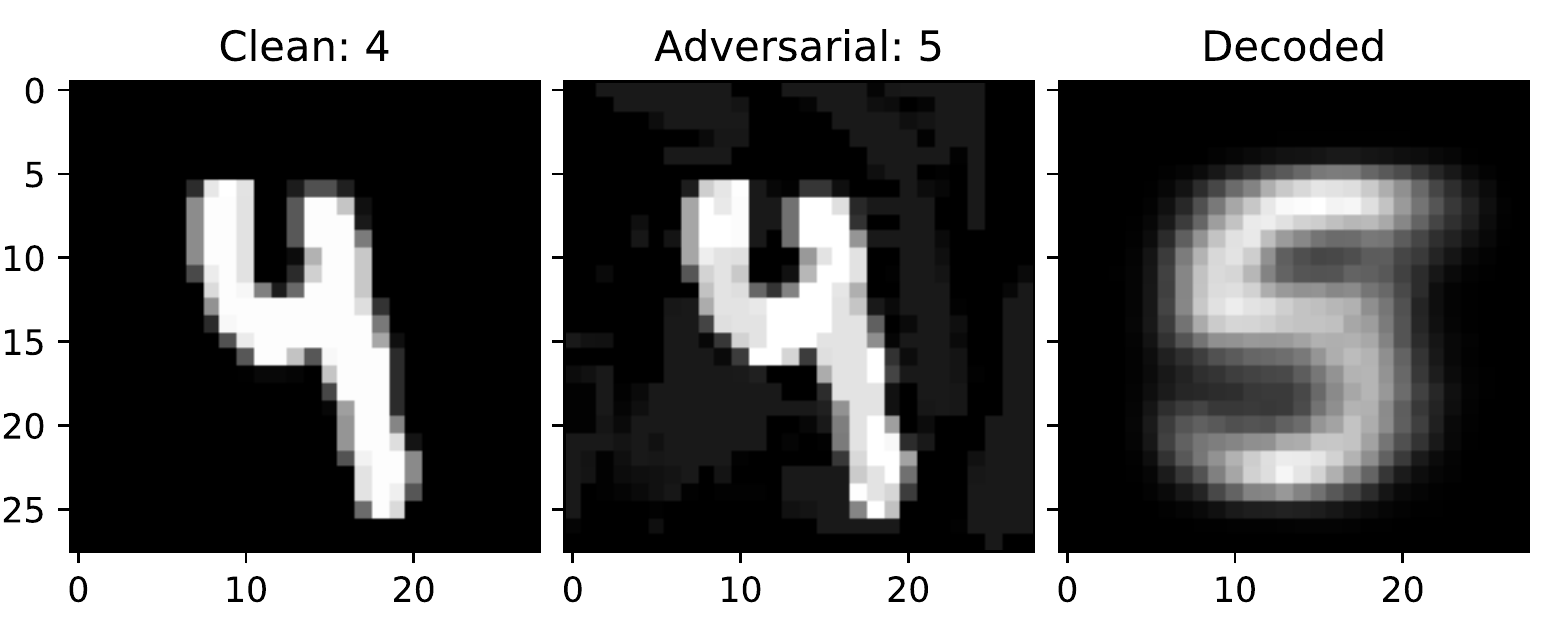}}
  	 	
  	   	\subfloat{%
      		 \includegraphics[width=1.0\columnwidth]{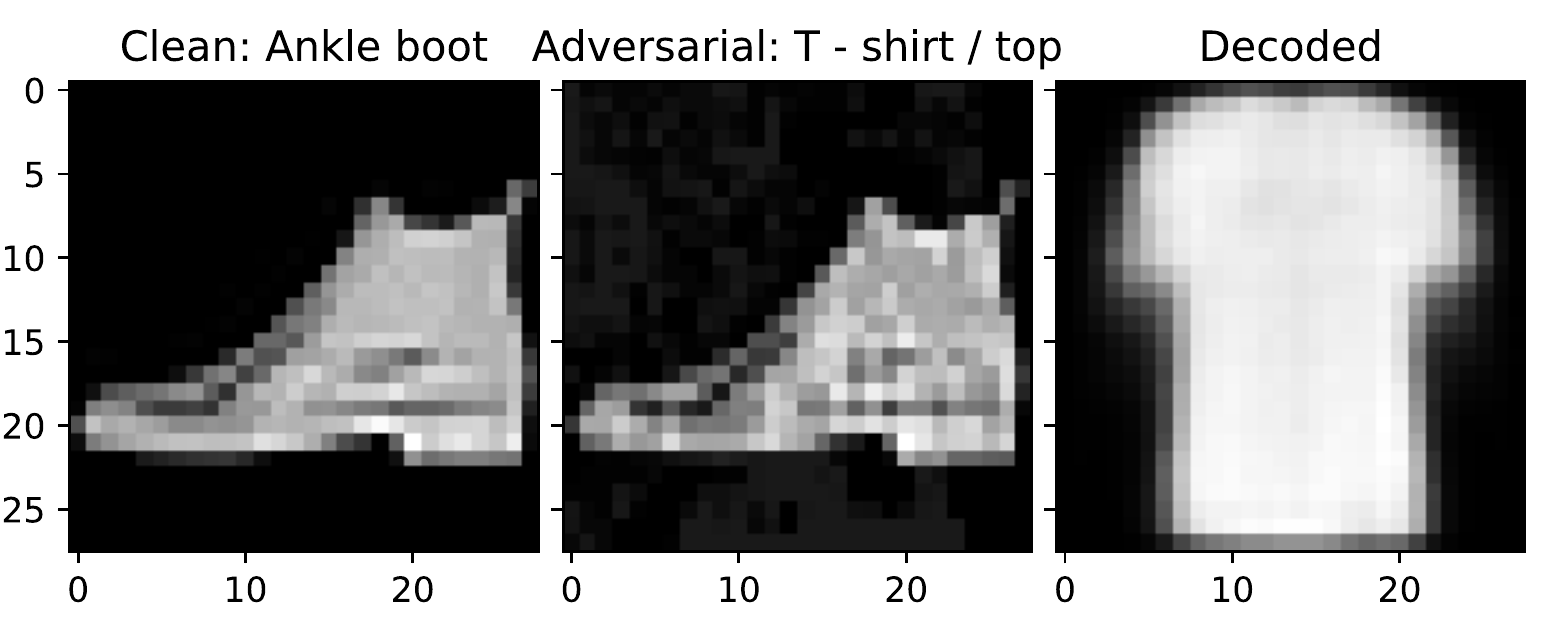}} 
      		 
  	 	
  	\subfloat{%
  	 	\includegraphics[width=1.0\columnwidth]{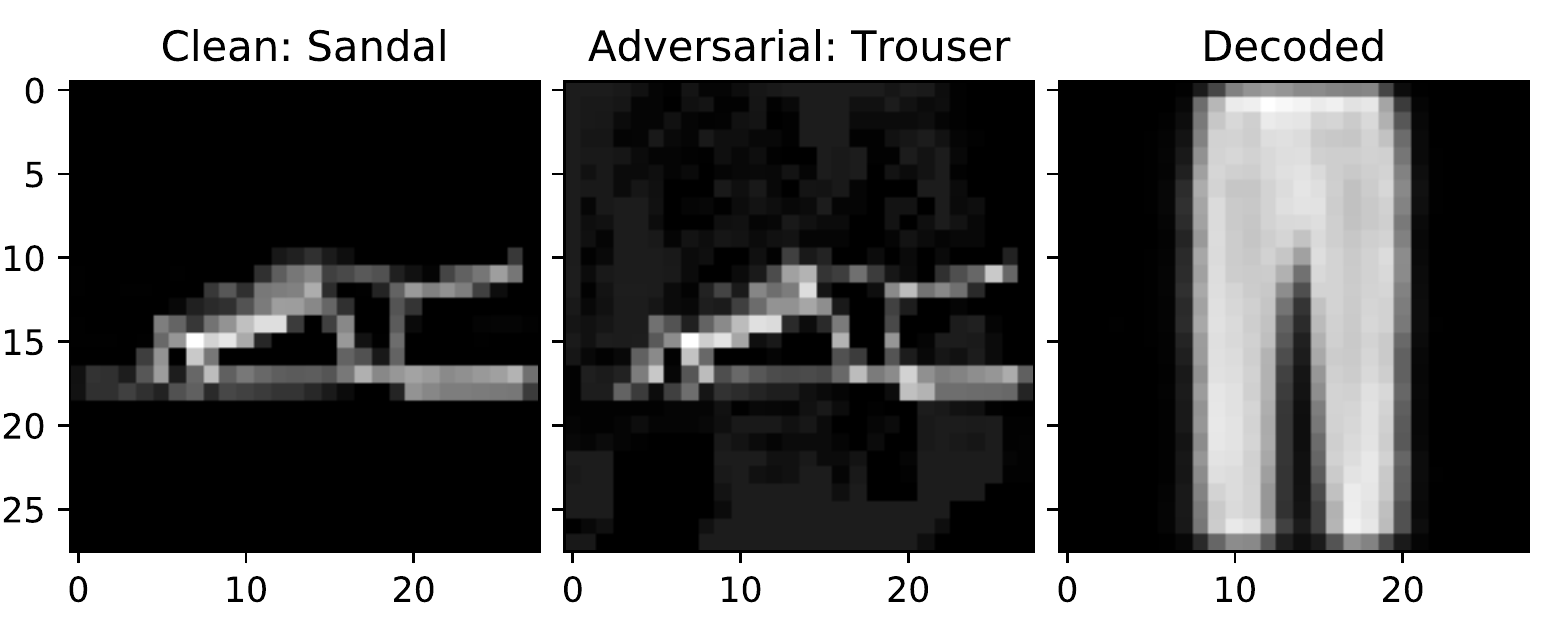}}\\
 \caption{Example of the effect of adversarial perturbations on the decoder network. It is noticeable how the reconstruction belongs to the class mistakenly chosen by the classifier.}
\label{fig:decoded} 
\end{figure}

We analyse here the effect of an adversarial perturbation on the decoded images of the autoencoder. Interestingly, from Figure \ref{fig:decoded} we notice that the decoder is fooled to reconstruct an image that seems to belong to the same incorrect class provided the classifier. For example in the first row the true class is 0, the mistaken class is 6 and the decoder produces an image similar to a 6. This suggests that a similar model may be used for other vision tasks such as conditional image generation and style transfer. We reserve this analysis as a future work.

%
%
\section{Conclusions} \label{concl}
In this paper we proposed an analysis of $\beta$-VAC in the presence of adversarial perturbations. We have shown that the model does not provide increased robustness to adversarial examples however it is able to detect them effectively thanks to the reconstruction error of the decoder network. Sparse regularization of the classifier does not help in reducing the effects of adversarial perturbations on the classification. We have shown that the decoder, when fed with an adversarial example, tend to reconstruct an image that belongs to the class mistakenly selected by the classifier. As a future work, we want to investigate deeper this aspect that may be exploited to perform conditional image generation and style transfer tasks. We also plan to extend these results to more complex vision datasets.

\bibliography{bib}             

\begin{thebibliography}{25}
\providecommand{\natexlab}[1]{#1}
\providecommand{\url}[1]{\texttt{#1}}
\providecommand{\urlprefix}{URL }
\expandafter\ifx\csname urlstyle\endcsname\relax
  \providecommand{\doi}[1]{doi:\discretionary{}{}{}#1}\else
  \providecommand{\doi}{doi:\discretionary{}{}{}\begingroup
  \urlstyle{rm}\Url}\fi

\bibitem[{Belkin et~al.(2018{\natexlab{a}})Belkin, Ma, and
  Mandal}]{belkin2018understand}
Belkin, M., Ma, S., and Mandal, S. (2018{\natexlab{a}}).
\newblock To understand deep learning we need to understand kernel learning.
\newblock \emph{arXiv:1802.01396}.

\bibitem[{Belkin et~al.(2018{\natexlab{b}})Belkin, Rakhlin, and
  Tsybakov}]{belkin2018does}
Belkin, M., Rakhlin, A., and Tsybakov, A.B. (2018{\natexlab{b}}).
\newblock Does data interpolation contradict statistical optimality?
\newblock \emph{arXiv:1806.09471}.

\bibitem[{Bishop(2006)}]{bishop2006pattern}
Bishop, C.M. (2006).
\newblock \emph{Pattern recognition and machine learning}.
\newblock springer.

\bibitem[{Botev et~al.(2010)Botev, Grotowski, Kroese et~al.}]{botev2010kernel}
Botev, Z.I., Grotowski, J.F., Kroese, D.P., et~al. (2010).
\newblock Kernel density estimation via diffusion.
\newblock \emph{The annals of Statistics}, 38(5), 2916--2957.

\bibitem[{Burgess et~al.(2018)Burgess, Higgins, Pal, Matthey, Watters,
  Desjardins, and Lerchner}]{burgess2018understanding}
Burgess, C.P., Higgins, I., Pal, A., Matthey, L., Watters, N., Desjardins, G.,
  and Lerchner, A. (2018).
\newblock Understanding disentangling in $\beta$-vae.
\newblock \emph{arXiv:1804.03599}.

\bibitem[{Feinman et~al.(2017)Feinman, Curtin, Shintre, and
  Gardner}]{feinman2017detecting}
Feinman, R., Curtin, R.R., Shintre, S., and Gardner, A.B. (2017).
\newblock Detecting adversarial samples from artifacts.
\newblock \emph{arXiv:1703.00410}.

\bibitem[{Finlay et~al.(2018)Finlay, Calder, Abbasi, and
  Oberman}]{finlay2018lipschitz}
Finlay, C., Calder, J., Abbasi, B., and Oberman, A. (2018).
\newblock Lipschitz regularized deep neural networks generalize and are
  adversarially robust.
\newblock \emph{arXiv:1808.09540}.

\bibitem[{Gal and Ghahramani(2016)}]{gal2016dropout}
Gal, Y. and Ghahramani, Z. (2016).
\newblock Dropout as a bayesian approximation: Representing model uncertainty
  in deep learning.
\newblock In \emph{ICLR}, 1050--1059.

\bibitem[{Gong et~al.(2017)Gong, Wang, and Ku}]{gong2017adversarial}
Gong, Z., Wang, W., and Ku, W.S. (2017).
\newblock Adversarial and clean data are not twins.
\newblock \emph{arXiv:1704.04960}.

\bibitem[{Goodfellow et~al.(2014)Goodfellow, Shlens, and
  Szegedy}]{goodfellow2014explaining}
Goodfellow, I.J., Shlens, J., and Szegedy, C. (2014).
\newblock Explaining and harnessing adversarial examples.
\newblock \emph{arXiv:1412.6572}.

\bibitem[{Gretton et~al.(2012)Gretton, Borgwardt, Rasch, Sch{\"o}lkopf, and
  Smola}]{gretton2012kernel}
Gretton, A., Borgwardt, K.M., Rasch, M.J., Sch{\"o}lkopf, B., and Smola, A.
  (2012).
\newblock A kernel two-sample test.
\newblock \emph{Journal of Machine Learning Research}, 13(Mar), 723--773.

\bibitem[{Grosse et~al.(2017)Grosse, Manoharan, Papernot, Backes, and
  McDaniel}]{grosse2017statistical}
Grosse, K., Manoharan, P., Papernot, N., Backes, M., and McDaniel, P. (2017).
\newblock On the (statistical) detection of adversarial examples.
\newblock \emph{arXiv:1702.06280}.

\bibitem[{Higgins et~al.(2017)Higgins, Matthey, Pal, Burgess, Glorot,
  Botvinick, Mohamed, and Lerchner}]{higgins2017beta}
Higgins, I., Matthey, L., Pal, A., Burgess, C., Glorot, X., Botvinick, M.,
  Mohamed, S., and Lerchner, A. (2017).
\newblock beta-vae: Learning basic visual concepts with a constrained
  variational framework.
\newblock \emph{ICLR}, 2(5), 6.

\bibitem[{Kingma and Welling(2013)}]{kingma2013auto}
Kingma, D.P. and Welling, M. (2013).
\newblock Auto-encoding variational bayes.
\newblock \emph{arXiv:1312.6114}.

\bibitem[{Li et~al.(2018)Li, Bradshaw, and Sharma}]{li2018generative}
Li, Y., Bradshaw, J., and Sharma, Y. (2018).
\newblock Are generative classifiers more robust to adversarial attacks?
\newblock \emph{arXiv:1802.06552}.

\bibitem[{Liu et~al.(2016)Liu, Chen, Liu, and Song}]{liu2016delving}
Liu, Y., Chen, X., Liu, C., and Song, D. (2016).
\newblock Delving into transferable adversarial examples and black-box attacks.
\newblock \emph{arXiv:1611.02770}.

\bibitem[{Madry et~al.(2017)Madry, Makelov, Schmidt, Tsipras, and
  Vladu}]{madry2017towards}
Madry, A., Makelov, A., Schmidt, L., Tsipras, D., and Vladu, A. (2017).
\newblock Towards deep learning models resistant to adversarial attacks.
\newblock \emph{arXiv:1706.06083}.

\bibitem[{Moosavi-Dezfooli et~al.(2019)Moosavi-Dezfooli, Fawzi, Uesato, and
  Frossard}]{moosavi2019robustness}
Moosavi-Dezfooli, S.M., Fawzi, A., Uesato, J., and Frossard, P. (2019).
\newblock Robustness via curvature regularization, and vice versa.
\newblock In \emph{Proceedings of the IEEE Conference on Computer Vision and
  Pattern Recognition}, 9078--9086.

\bibitem[{Nair and Hinton(2010)}]{nair2010rectified}
Nair, V. and Hinton, G.E. (2010).
\newblock Rectified linear units improve restricted boltzmann machines.
\newblock In \emph{Proceedings of the 27th international conference on machine
  learning (ICML-10)}, 807--814.

\bibitem[{Ross and Doshi-Velez(2018)}]{ross2018improving}
Ross, A.S. and Doshi-Velez, F. (2018).
\newblock Improving the adversarial robustness and interpretability of deep
  neural networks by regularizing their input gradients.
\newblock In \emph{Thirty-second AAAI conf. on artificial intelligence}.

\bibitem[{Shaham et~al.(2018)Shaham, Yamada, and
  Negahban}]{shaham2018understanding}
Shaham, U., Yamada, Y., and Negahban, S. (2018).
\newblock Understanding adversarial training: Increasing local stability of
  supervised models through robust optimization.
\newblock \emph{Neurocomputing}, 307, 195--204.

\bibitem[{Springenberg et~al.(2014)Springenberg, Dosovitskiy, Brox, and
  Riedmiller}]{springenberg2014striving}
Springenberg, J.T., Dosovitskiy, A., Brox, T., and Riedmiller, M. (2014).
\newblock Striving for simplicity: The all convolutional net.
\newblock \emph{arXiv:1412.6806}.

\bibitem[{Szegedy et~al.(2013)Szegedy, Zaremba, Sutskever, Bruna, Erhan,
  Goodfellow, and Fergus}]{szegedy2013intriguing}
Szegedy, C., Zaremba, W., Sutskever, I., Bruna, J., Erhan, D., Goodfellow, I.,
  and Fergus, R. (2013).
\newblock Intriguing properties of neural networks.
\newblock \emph{arXiv:1312.6199}.

\bibitem[{Terzi et~al.(2020)Terzi, Susto, and Chaudhari}]{terzi2019directional}
Terzi, M., Susto, G.A., and Chaudhari, P. (2020).
\newblock Directional adversarial training for cost sensitive deep learning
  classification applications.
\newblock \emph{Engineering Applications of Artificial Intelligence}, 91,
  103550.

\bibitem[{Tsipras et~al.(2018)Tsipras, Santurkar, Engstrom, Turner, and
  Madry}]{tsipras2018robustness}
Tsipras, D., Santurkar, S., Engstrom, L., Turner, A., and Madry, A. (2018).
\newblock Robustness may be at odds with accuracy.
\newblock \emph{arXiv:1805.12152}.

\end{thebibliography}
                                                   







\end{document}